%% file: main.tex
\PassOptionsToPackage{dvipsnames,table}{xcolor}
\documentclass[runningheads]{llncs}
\usepackage[T1]{fontenc}
\usepackage{graphicx, verbatim}
\usepackage{multirow}
\usepackage{amsmath}

\usepackage{array}
\usepackage{amssymb}
\usepackage{cite}
\usepackage{subcaption}
\usepackage[colorlinks=false]{hyperref}
\usepackage{cleveref}
\usepackage{enumitem}
\usepackage{xcolor}
\usepackage{booktabs}
\def\etal{\emph{et al.}}
\hypersetup{
    colorlinks,
    linkcolor={red!80!black},
    citecolor={blue!80!black},
    urlcolor={}
}
\usepackage{float}
\usepackage[table]{xcolor} % Enables cell coloring
\usepackage{makecell}
\usepackage{tabularx, makecell, booktabs}
\newcommand\norm[1]{\left\lVert#1\right\rVert}
\usepackage{orcidlink}
\def\discintname{Disclosure of Interests.}

\begin{document}
\title{Out-of-Distribution Detection in Medical Imaging via Diffusion Trajectories}
\titlerunning{OODd in Medical Imaging via Diffusion Trajectories}

% \begin{comment}
%% Removed for anonymized MICCAI 2025 submission
\author{
Lemar Abdi~\orcidlink{0009-0003-9301-1613} \and
Francisco Caetano~\orcidlink{0000-0002-6069-6084} \and
Amaan Valiuddin~\orcidlink{0009-0005-2856-5841} \and
Christiaan Viviers~\orcidlink{0000-0001-6455-0288} \and Hamdi Joudeh~\orcidlink{0000-0002-7162-0325} \and
Fons van der Sommen~\orcidlink{0000-0002-3593-2356}
}
\authorrunning{L. Abdi et al.}
\institute{Eindhoven University of Technology, Eindhoven 5612 AZ, The Netherlands
\email{l.abdi@tue.nl}}
% \end{comment}

% \author{Anonymized Authors}  %% Added for anonymized MICCAI 2025 submission
% \authorrunning{Anonymized Author et al.}
% \institute{Anonymized Affiliations \\
%     \email{email@anonymized.com}}

% meta comment

\maketitle              % typeset the header of the contribution
\begin{abstract}
In medical imaging, unsupervised out-of-distribution~(OOD) detection offers an attractive approach for identifying pathological cases with extremely low incidence rates. In contrast to supervised methods, OOD-based approaches function without labels and are inherently robust to data imbalances. Current generative approaches often rely on likelihood estimation or reconstruction error, but these methods can be computationally expensive, unreliable, and require retraining if the inlier data changes. These limitations hinder their ability to distinguish nominal from anomalous inputs efficiently, consistently, and robustly. We propose a reconstruction-free OOD detection method that leverages the forward diffusion trajectories of a Stein score-based denoising diffusion model~(SBDDM). By capturing trajectory curvature via the estimated Stein score, our approach enables accurate anomaly scoring with only five diffusion steps. A single SBDDM pre-trained on a large, semantically aligned medical dataset generalizes effectively across multiple Near-OOD and Far-OOD benchmarks, achieving state-of-the-art performance while drastically reducing computational cost during inference. Compared to existing methods, SBDDM achieves a relative improvement of up to 10.43\% and 18.10\% for Near-OOD and Far-OOD detection, making it a practical building block for real-time, reliable computer-aided diagnosis.
\end{abstract}

\input{chapters/1-Intro}

\input{chapters/2-Background}

\input{chapters/3-Method}

\input{chapters/4-Results}

\input{chapters/5-Conclusion}

\subsubsection{\protect\discintname}
The authors have no competing interests to declare that are relevant to the content of this article. 

% \clearpage

\bibliographystyle{splncs04}
\bibliography{ref}

\end{document}

%% file: chapters/1-Intro.tex
\section{Introduction}

Out-of-distribution~(OOD) detection is a critical task in medical imaging, where real-world deployments often face data that diverges from the training distribution due to rare pathologies, acquisition variability, or previously unseen patient demographics. These challenges, which are aggravated by low prevalence, severe class imbalance, and the high cost of expert annotation, undermine the effectiveness of supervised approaches~\cite{li2016improving}. In response, unsupervised OOD detection has emerged as a promising alternative, which is solely trained on the readily available, typically more abundant non-pathological cases.

Unsupervised OOD approaches typically operate by modeling the inlier distribution of training data, explicitly through likelihood estimation~\cite{zhao2022ae, valiuddin2022efficient}, or implicitly through distance metrics at the pixel or feature level~\cite{denouden2018improving, li2023rethinking}. 
% Density-based methods aim to assign lower likelihoods to anomalous inputs, but in practice, deep generative models often assign high likelihoods to OOD samples~\cite{nalisnick_deep_2019, choi_waic_2019, kirichenko_why_2020}. 
Reconstruction-based methods attempt to flag anomalies based on reconstruction error, assuming that models trained on inlier data will fail to accurately reconstruct unseen or pathological structures. Among these, Denoising Diffusion Models~(DDMs) have shown strong reconstruction performance in medical imaging~\cite{wolleb2022diffusion,naval2024ensembled,behrendt2024leveraging}, but suffer from key limitations: they require multiple of sampling steps in the forward and reverse process, are sensitive to noise level choices, and can also inadvertently reconstruct anomalous inputs faithfully---leading to false negatives. Moreover, conditioning mechanisms used during reconstruction may introduce dataset-specific biases. To overcome these challenges, recent works~\cite{heng2024out, sakai2025reconstruction} have proposed reconstruction-free OOD detection using the \emph{forward}~(noising) process of DDMs. These methods hypothesize that the diffusion trajectory of anomalous inputs deviates from that of nominal data, allowing effective OOD detection with significantly fewer model evaluations.

In this work, we extend forward diffusion-based OOD detection to the medical domain by leveraging the trajectories of DDMs, as realized through the estimated Stein scores. Our method avoids the limitations of reconstruction-based approaches and achieves strong performance, using as few as five diffusion steps. We further show that the learned score function generalizes across datasets, enabling OOD detection on multiple medical benchmarks with a single pre-trained model. The proposed approach outperforms state-of-the-art~(SOTA) methods across twelve diverse medical imaging benchmarks in both Near-OOD and Far-OOD settings.

%% file: chapters/2-Background.tex
\section{Related Work} 

\subsection{Score-based Diffusion Models}

Score-based diffusion models gradually perturb data $\mathbf{x}_0$ from data distribution $p_0$ into a simple prior distribution $p_T$, typically a standard Gaussian, through a stochastic differential equation~(SDE). The forward diffusion process $\{\mathbf{x}_t\}_{t=0}^T$ is governed by the SDE:
\begin{equation}\label{eq:sde}
\text{d}\mathbf{x} = \mathbf{f}(\mathbf{x}, t) \text{d}t + g(t)\text{d}\mathbf{w},
\end{equation}
where $\mathbf{f}(\mathbf{x}, t)$ denotes the drift term, $g(t)$ is a scalar diffusion coefficient, and $\mathbf{w}$ is a standard Wiener process.
Sampling from the learned data distribution involves reversing this diffusion process. The reverse-time SDE is given by:
\begin{equation}
\text{d}\mathbf{x} = \left[ \mathbf{f}(\mathbf{x}, t) - g(t)^2 \nabla_\mathbf{x} \log p_t(\mathbf{x}) \right]\text{d}t + g(t)\text{d}\bar{\mathbf{w}},
\end{equation}
where $\nabla_\mathbf{x} \log p_t(\mathbf{x})$ is the \textit{Stein score} (or score function), and $\bar{\mathbf{w}}$ denotes the reverse Brownian motion over $[T, 0]$.
Song~\etal~\cite{song2021scorebased} showed that this stochastic process has a deterministic counterpart, known as the probability flow ODE, which yields the same marginal distributions as the original SDE. This formulation allows the diffusion process to be expressed as:
\begin{equation}\label{eq:pf}
\text{d}\mathbf{x} = \left[ \mathbf{f}(\mathbf{x}, t) - \frac{1}{2}g(t)^2 \nabla_\mathbf{x} \log p_t(\mathbf{x}) \right]\text{d}t.
\end{equation}
Both the reverse-time SDE and the probability flow ODE require an accurate estimate of the score function. This is typically achieved by training a neural network $\boldsymbol{\epsilon}_\theta(\mathbf{x}, t)$ to approximate the Stein score function, using the score matching objective~\cite{vincent2011connection}:
\begin{equation}
\min _\theta \mathbb{E}_{t \sim \mathcal{U}[0,T]}  \left\{ \mathbb{E}_{\mathbf{x}_0 \sim p_0(\mathbf{x}_0)} \mathbb{E}_{\mathbf{x}_t \sim p_{0t}(\mathbf{x}_t | \mathbf{x}_0)}\left[\left\|\boldsymbol{\epsilon}_\theta\left(\mathbf{x}_t, t\right)-\boldsymbol{\epsilon}\right\|_2^2\right] \right\},
\end{equation}
where the target denoising direction is defined as $\boldsymbol{\epsilon} = -\sigma_t \nabla_{\mathbf{x}_t} \log p_{0t}(\mathbf{x}_t \mid \mathbf{x}_0)$ and $p_{0t}(\mathbf{x}_t \mid \mathbf{x}_0) = \mathcal{N}(\sqrt{\bar{\alpha}_t} \mathbf{x}_0, \sigma_t^2 \mathbf{I})$, following the DDPM formulation~\cite{ho2020denoising} for the fixed time-dependent variances $\sigma_t^2 = (1 - \bar{\alpha}_{t-1})/(1 - \bar{\alpha}_t)$.

\subsection{Unsupervised OOD Detection}

Given an in-distribution~(ID) dataset sampled from $p_{\text{ID}}$, the goal of unsupervised OOD detection is to identify test-time samples that lie outside high-probability mass regions of $p_{\text{ID}}$, without requiring labels or prior knowledge of OOD samples. Existing approaches generally fall into three categories, described as follows. (1)~Distance-based methods (also known as feature-based methods) compare test samples with training data in a learned embedding space, using metrics such as cosine similarity or Mahalanobis distance. These approaches rely on the assumption that OOD samples are farther from the ID feature manifold. Although effective, they often require storing training embeddings or class prototypes in memory~\cite{yang2024generalizedood}. (2)~Density-based methods attempt to estimate $\log p_{\text{ID}}(\mathbf{x})$ directly. Normalizing Flows~(NFs) enable exact likelihood computation, while VAEs and diffusion models yield tractable bounds on the likelihood. Despite their theoretical motivation, deep generative models often assign higher likelihoods to OOD inputs, undermining their effectiveness~\cite{nalisnick_deep_2019, choi_waic_2019, kirichenko_why_2020}. (3)~Reconstruction-based methods assume that OOD samples will be poorly reconstructed by generative models trained on ID data, which can be quantified in pixel-space or feature-space. However, these methods often fail in complex settings---such as medical imaging---where anomalous samples can still be faithfully reconstructed~\cite{denouden2018improving, zhou2022rethinking}.
\\
\par
\textbf{OOD Detection with Diffusion Models:} Diffusion models have gained traction in the context of OOD detection, primarily through reconstruction-based strategies. These methods typically employ class-conditioned DDMs, where the model learns to reconstruct inputs conditioned on non-pathological (often \emph{healthy}) class labels~\cite{wolleb2022diffusion, mousakhan2024anomaly}. Recent methods improve robustness by ensembling multiple reconstructions for distance computations or conditioning~\cite{naval2024ensembled, behrendt2024leveraging}. Nevertheless, these methods require many sampling steps to reconstruct an image---making real-time deployment less practical in clinical settings. Density-based approaches using DDMs are also feasible. The probability flow ODE (\cref{eq:pf}) allows for the exact likelihood computation under score-based diffusion models.  However, this approach still suffers from the same limitations seen in other likelihood-based approaches~\cite{graham2023denoising}. More recently, reconstruction-free strategies have emerged that take advantage of the forward (noising) process of diffusion models~\cite{heng2024out, sakai2025reconstruction}. These methods hypothesize that the forward trajectory of an anomalous sample deviates significantly from that of inlier data. In particular, they achieve strong performance while requiring a fewer number of function evaluations~(NFEs), highlighting their potential for clinical applications.

%% file: chapters/3-Method.tex
\section{Method}

\subsection{Stein Score DDM OOD Detection}
\begin{figure}
    \centering
    \includegraphics[width=\linewidth]{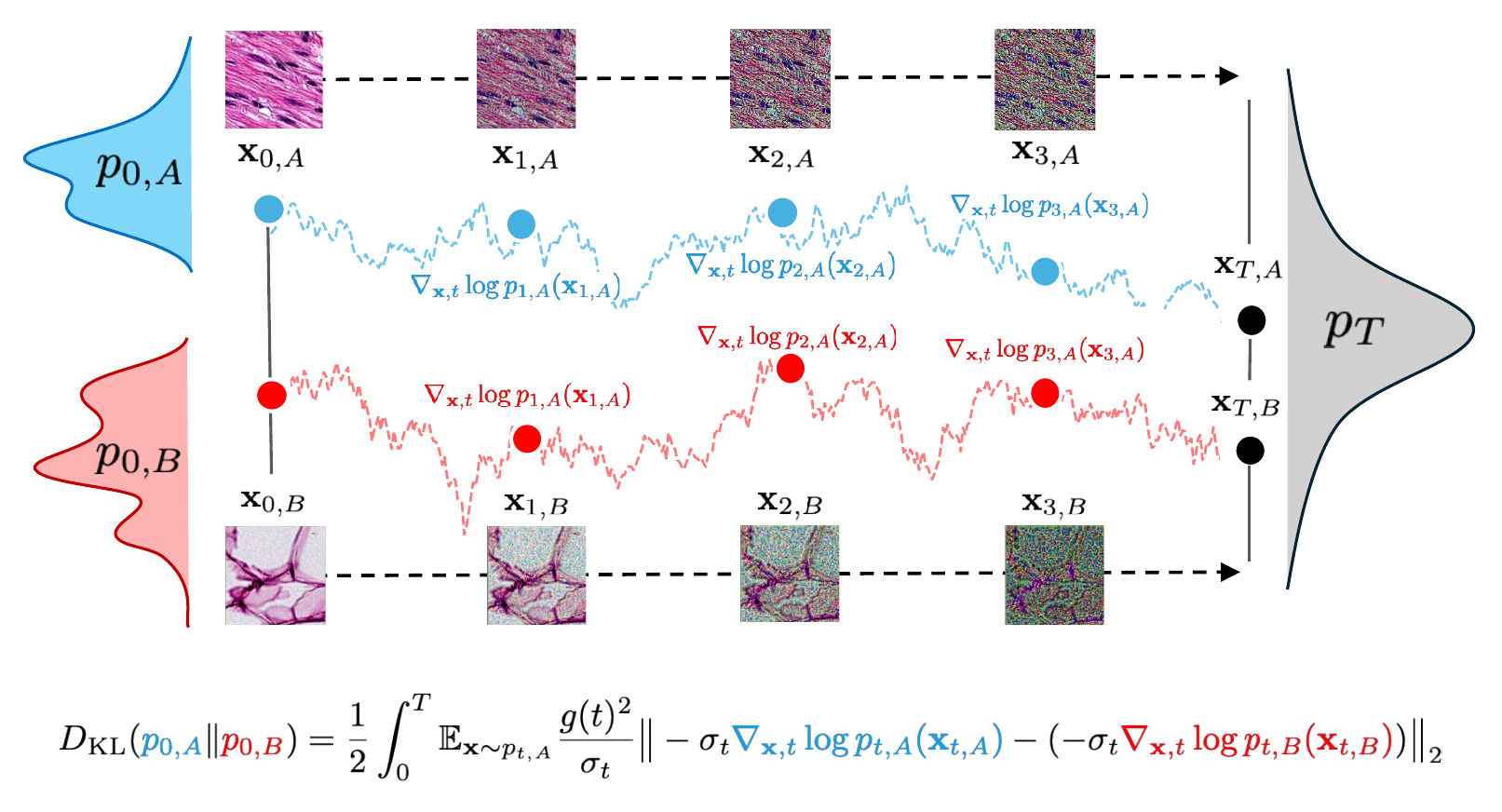}
    \caption{Distinct data distributions induce different curvatures along the forward diffusion path. The estimated Stein score characterizes whether a given trajectory conforms to the curvature of the ID density, thereby enabling a reconstruction-free OOD detection approach with unconditional DDMs.}
    \label{fig:method}
\end{figure}

Given two distinct data distributions $p_{0,A}$ and $p_{0,B}$, the intermediate marginals $p_{t,A}$ and $p_{t,B}$ during the diffusion process are governed by the deterministic probability flow ODE of~\Cref{eq:pf}. These marginals along the forward-time ODE path---i.e., from sample to noise---differ depending on the data distribution. Under regularity conditions, the divergence between these two distributions is equivalent to the squared difference between their Stein scores, as shown in~\cite{heng2024out}:
\begin{equation}\label{eq:dkl}
    D_\mathrm{KL}(p_{0,A} \| p_{0,B}) = \frac{1}{2} \int_0^T \mathbb{E}_{\mathbf{x} \sim p_{t,A}} 
    \frac{g(t)^2}{\sigma_t} \big\| \boldsymbol{\epsilon}_A
-\boldsymbol{\epsilon}_B \big\|_2 
    + \underbrace{D_\mathrm{KL}(p_{T,A} \| p_{T,B})}_{=0} \:,
\end{equation}
where $\boldsymbol{\epsilon}_{i} \approx -\sigma_t \nabla_{\mathbf{x},t} \log p_{t,i}(\mathbf{x}_{t,i})$ denotes the estimated Stein score at time $t$.
The estimated Stein score reflects the local gradient of the marginals in the diffusion process, effectively capturing the direction and rate of change of the forward trajectory. Its evolution over time---i.e., the variation of the score along the diffusion path---captures curvature, which further distinguishes \emph{nominal} from \emph{anomalous} samples, as illustrated in~\Cref{fig:method}. Using this, the anomaly score $s(\cdot)$ is defined for a test sample $\mathbf{x}_0^*$ as:
\begin{equation}\label{eq:testscore}
    s(\mathbf{x}_0^*) = \norm{\sum_{t=1}^S-\sigma_t\nabla_{\mathbf{x},t}\log p_t({\mathbf{x}_t^*})}^p + \norm{\sum_{t=1}^S\frac{\partial}{\partial t}\Bigl(-\sigma_t\nabla_{\mathbf{x},t}\log p_t({\mathbf{x}_t^*})\Big)}^p,
\end{equation}
where $S$ denotes the number of forward-time diffusion steps, and $\norm{\cdot}_p$ the $p$-norm. 
Similarly to Heng~\etal~\cite{heng2024out}, a Kernel Density Estimation (KDE) with a Gaussian kernel is fitted to the scores of the inlier validation samples $ s(\mathbf{x}^{\text{val}}_0)$ of each benchmark. At test time, a sample $\mathbf{x}_0^*$ is marked as OOD if the likelihood of its score $s(\mathbf{x}_0^*)$ under the KDE is less than $s(\mathbf{x}^{\text{val}}_0)$. 

We follow OpenAI's \texttt{improved diffusion}\footnote{\url{https://github.com/openai/improved-diffusion}} codebase and architectural design. The models are trained using a cosine noise schedule, uniform time sampling, a learning rate of 1e-4, and a batch size of 64. For both training and inference, we use the DDIM sampler~\cite{song2021denoising}. During testing, the Stein scores in~\Cref{eq:testscore} are computed using the $\boldsymbol{\epsilon}$ output of the \texttt{ddim\_reverse\_sample} function. We fix $S = 5$, $p = 3$, and compute the time derivative using finite difference.

Unlike many existing methods, this approach does not require retraining on each inlier dataset, allowing application across multiple datasets with a single model. The proposed unconditional DDM relies on a well-trained score estimator and a KDE fitted to the validation data of the inlier dataset, which ensures that~\Cref{eq:dkl} remains valid regardless of the dataset used to train $\boldsymbol{\epsilon}_\theta$. However, the choice of training data remains important: the model must learn features that generalize well within the medical domain. We empirically analyze this dependency and its impact on cross-dataset generalization in~\Cref{sec:ablations}.

\subsection{Medical OOD Benchmark} 

We adopt the OpenOOD framework~\cite{openoodyang2022} in the medical domain for unsupervised image-level OOD detection across the Near-OOD and Far-OOD regimes. We adopt five MedMNIST datasets~\cite{medmnistyang2023} and the COVID-19 dataset~\cite{chowdhury2020can_covid, rahman2021exploring_covid} to cover a spectrum of semantic and modality shifts. We define \underline{\emph{Near-OOD}} as a semantic shift within the same dataset: samples from classes not seen during training but drawn from the same support. Specifically, we split each dataset by aggregating training classes into majority/minority partitions and treat the excluded subset as OOD. For binary-class datasets, this corresponds to `healthy' vs. `pathological'; for multi-class datasets, it involves excluding minority or semantically disjoint classes. This partitioning aligns with the clinically relevant setting of detecting the pathologies absent in the training data. Our splitting protocol follows Narayanaswamy~\textit{et al.}~\cite{narayanaswamy2023exploring} to ensure a consistent comparison. In contrast, \underline{\emph{Far-OOD}} corresponds to modality shifts across datasets. In this setting, one dataset is designated as ID, and all the remaining datasets serve as sources of OOD data. All images are resized to 128$\times$128 pixels.

\subsection{Baselines} 
We compare our proposed method with a range of state-of-the-art~(SOTA) approaches, including both supervised and unsupervised methods:

\textbf{Classification-based}: We include two semi-supervised approaches with outlier-exposure---VOS++ and NDA++~\cite{du2022towards,hendrycks*2020augmix,sinha2021negative,narayanaswamy2023exploring}---which are trained on additional synthetic samples compared to their vanilla models. G-ODIN~\cite{hsu2020generalized} is an unsupervised confidence-based method that combines temperature scaling and input perturbation to improve softmax OOD discrimination.

\textbf{Density-based}: Mubarka \etal~\cite{narayanaswamy2023exploring} propose an energy-based model trained on additional synthetic samples, achieving the SOTA OOD performance in MedMNIST prior to this work. We also evaluate GLOW~\cite{kingma2018glow}, a normalizing flow (NF) model evaluated on the exact data log-likelihood, and Typ-Score~\cite{abdi2024typicality, viviers2025can}, an NF method that quantifies sample typicality using the Stein score of the data log-likelihood.

\textbf{Reconstruction-based}: We include a DDM conditioned on “healthy” samples~\cite{graham2023denoising}, using a DDIM~\cite{song2021denoising} sampler. For anomaly scoring, we evaluate both a combined feature-level LPIPS and pixel-level L2 error, as well as LPIPS alone, denoted as cDDM~(L).

\textbf{Forward diffusion-based}: Sakai~\etal~\cite{sakai2025reconstruction} propose a reconstruction-free diffusion method, which similarly to SBDDM(-P), obtains anomaly scores using solely the forward-time diffusion path. We follow the same hyperparameter settings, with the model  conditioned on healthy inlier images.
\begin{table*}[hbt]
    \centering
    \scriptsize 
    \caption{AUROC scores (Near-OOD / Far-OOD). 
    Color-coded results are obtained by SBDDM-P trained on \textcolor{blue}{PathMNIST} or \textcolor{orange}{TissueMNIST}. ($^*$) denotes re-implementation.}
    \vspace{.25cm}
    \begin{tabularx}{\textwidth}{>{\raggedright\arraybackslash}X
                                 |>{\centering\arraybackslash}X
                                 |>{\centering\arraybackslash}X
                                 |>{\centering\arraybackslash}X
                                 |>{\centering\arraybackslash}X
                                 |>{\centering\arraybackslash}X
                                 |>{\centering\arraybackslash}X
                                 |>{\centering\arraybackslash}X}
        \toprule
        \textbf{Method} & \makecell{\textbf{Blood-} \\ \textbf{MNIST}} & \makecell{\textbf{Derma-} \\ \textbf{MNIST}} & \makecell{\textbf{Path-} \\ \textbf{MNIST}} & \makecell{\textbf{Tissue-} \\ \textbf{MNIST}} & \makecell{\textbf{Pneu-} \\ \textbf{monia}} & \textbf{Covid-19} & \textbf{Avg.} \\
        \hline
        G-ODIN       & 53.9 / 88.7 & 69.3 / 85.3 & 51.7 / 84.4 & 55.2 / 82.7& -     & -     & 57.5 / 85.3 \\
        VOS++        & 38.2 / 84.2 & 72.9 / 85.3 & 71.7 / 71.0 & 28.0 / 60.2 & -     & -     & 52.7 / 75.2 \\
        NDA++        & 53.5 / 95.8 & 69.9 / 80.0 & 57.0 / 61.1 & 58.8 / 70.2 & -     & -     & 59.8 / 76.8 \\
        Mubarka      & \underline{89.1} / \underline{99.7} & \underline{75.5} / 96.6 & 71.2 / \textbf{98.9} & \textbf{83.4} / \underline{96.6} & - & - & \underline{79.8} / \underline{97.9} \\
        GLOW$^*$         & 33.9 / 45.4 & 53.9 / 23.7 & 64.0 / 33.1 & 65.0 / 50.0 & 55.7 / 51.2 & 65.4 / 09.0 & 56.3 / 35.4 \\
        Typ-score$^*$    & 55.7 / 91.4 & 70.8 / \textbf{100} & 82.2 / 81.7 & \underline{70.7} / 95.3 & \textbf{90.4} / 91.7 & \underline{93.7} / 57.3 & 77.3 / 86.2 \\
        cDDM~(L)$^*$        & 71.1 / 78.7 & 74.4 / 73.2 & 81.2 / 83.0 & 44.3 / 63.9 & 81.5 / 86.2 & 69.9 / {82.3} & 70.4 / 77.9 \\
        cDDM$^*$         & 84.2 / 86.0 & 74.8 / 66.0 & {83.4} / 91.4 & 62.4 / 80.0 & 82.6 / 72.8 &  67.2 / 68.2 &  75.8 / 77.4 \\
        Sakai$^*$         & 56.5 / 77.4 & 72.0 / 82.5 & \underline{91.5} / \underline{97.4} & 58.0 / 86.0 & 82.2 / {87.9} &  \textbf{95.6} / 87.7 &  76.0 / 86.4 \\
        \midrule
        SBDDM         &  88.6 / \textbf{100} & 67.9 / \underline{99.9}& \textbf{92.1} / {96.6} & 64.6 / \textbf{99.9} & 73.8 / \underline{96} & 80 / \textbf{97.2} & 77.8 / \textbf{98.3}
        \\
        SBDDM-P         & \cellcolor{blue!20} \textbf{89.6} / 95.6 & \cellcolor{blue!20} \textbf{77.3} / 99.1 & \cellcolor{blue!20} \textbf{92.1} / {96.6} & \cellcolor{orange!20} 64.6 / \textbf{99.9} & \cellcolor{orange!20} \underline{86.8} / \textbf{100} & \cellcolor{orange!20} {89.3} / 70.7 & \textbf{83.3} / 93.7\\
        \bottomrule
    \end{tabularx}
    \label{tab:ood_comparison}
\end{table*}

%% file: chapters/4-Results.tex
\section{Results and Discussion}

\subsection{OOD Benchmark Performance}\label{sec:mainresults}

We present the OOD detection performance of the proposed Stein score-based DDM~(SBDDM) method against several baselines in~\Cref{tab:ood_comparison}, using the Area Under the Receiver Operating Characteristic curve~(AUROC) as the primary metric. Far-OOD AUROC results are averaged across all OOD datasets per benchmark. The classification-based baseline results are sourced from~\cite{narayanaswamy2023exploring}.
We report two variants of our method: SBDDM, which is retrained per inlier dataset (analogous to the baselines), and SBDDM-P, which uses a single model pre-trained on a large medical dataset (PathMNIST for RGB benchmarks, TissueMNIST for grayscale) across all benchmarks. Interestingly, SBDDM-P often outperforms retrained counterparts, underscoring the benefits of domain-consistent Stein score estimation; we further analyze this effect in~\Cref{sec:ablations}. Among the baselines, Typ-score achieves the best Far-OOD result on DermaMNIST and the best Near-OOD result on Covid-19. Mubarka~\etal~\cite{narayanaswamy2023exploring} achieve the best Far-OOD performance on PathMNIST and the best Near-OOD performance on TissueMNIST, while Sakai~\etal~\cite{sakai2025reconstruction} achieve the best Near-OOD performance on Covid-19. In the remaining seven benchmarks, the proposed method outperforms all baselines across both Near-OOD and Far-OOD settings. Additionally, when running on an NVIDIA H100 GPU, SBDDM has a latency of 55~ms~(18.2 FPS) using 5 NFEs, compared to 940~ms (1.1 FPS) with 100 NFEs for the reconstruction-based cDDM baseline.
\begin{table}[ht]
    \centering
    \caption{Effect of training distribution  $p_{\text{train}}$ of the SBDMM on Near-OOD performance.}
    \resizebox{0.55\columnwidth}{!}{%
    \begin{tabular}{l | c | c }
        \toprule
        {$p_{\text{train}}$} & \: \textbf{BloodMNIST} \: & \: \textbf{Pneumonia} \:\\
        \midrule
        ImageNet & 0.597  & 0.456 \\
        BloodMNIST & 0.886  & 0.738 \\
        PathMNIST &  0.896 & 0.868\\
        MedMNIST RGB & 0.775 & - \\ 
        MedMNIST Gray & - & 0.597\\
        \bottomrule
    \end{tabular}%
    }
    \label{tab:ablation1}
\end{table}
\subsection{Ablations}\label{sec:ablations}

We evaluate the impact of the training distribution $p_{\text{train}}$ on OOD detection performance in~\Cref{tab:ablation1}. Using a score-based DDM trained on ImageNet results in significantly lower performance, confirming that generic natural image statistics are insufficient to capture meaningful diffusion patterns in the medical domain. Training on the same dataset used for ID testing---such as BloodMNIST---yields strong OOD detection performance, as expected. Surprisingly, training on PathMNIST, a larger medical dataset with distinct semantics, leads to even better results. However, aggregating all the RGB datasets from MedMNIST into one large training set (MedMNIST RGB) and combining the grayscale datasets into another (MedMNIST Grey) introduces inconsistencies and often degrades performance. This may stem from the distortion of transport paths and flattening of curvatures that are informative for detecting near-OOD samples. These findings suggest that $p_{\text{train}}$ should be drawn from a sufficiently large and semantically consistent dataset, ensuring that the score estimation generalizes to other benchmarks within the same domain.
\begin{figure}[hbt]
    \centering
    \includegraphics[width=0.7\linewidth]{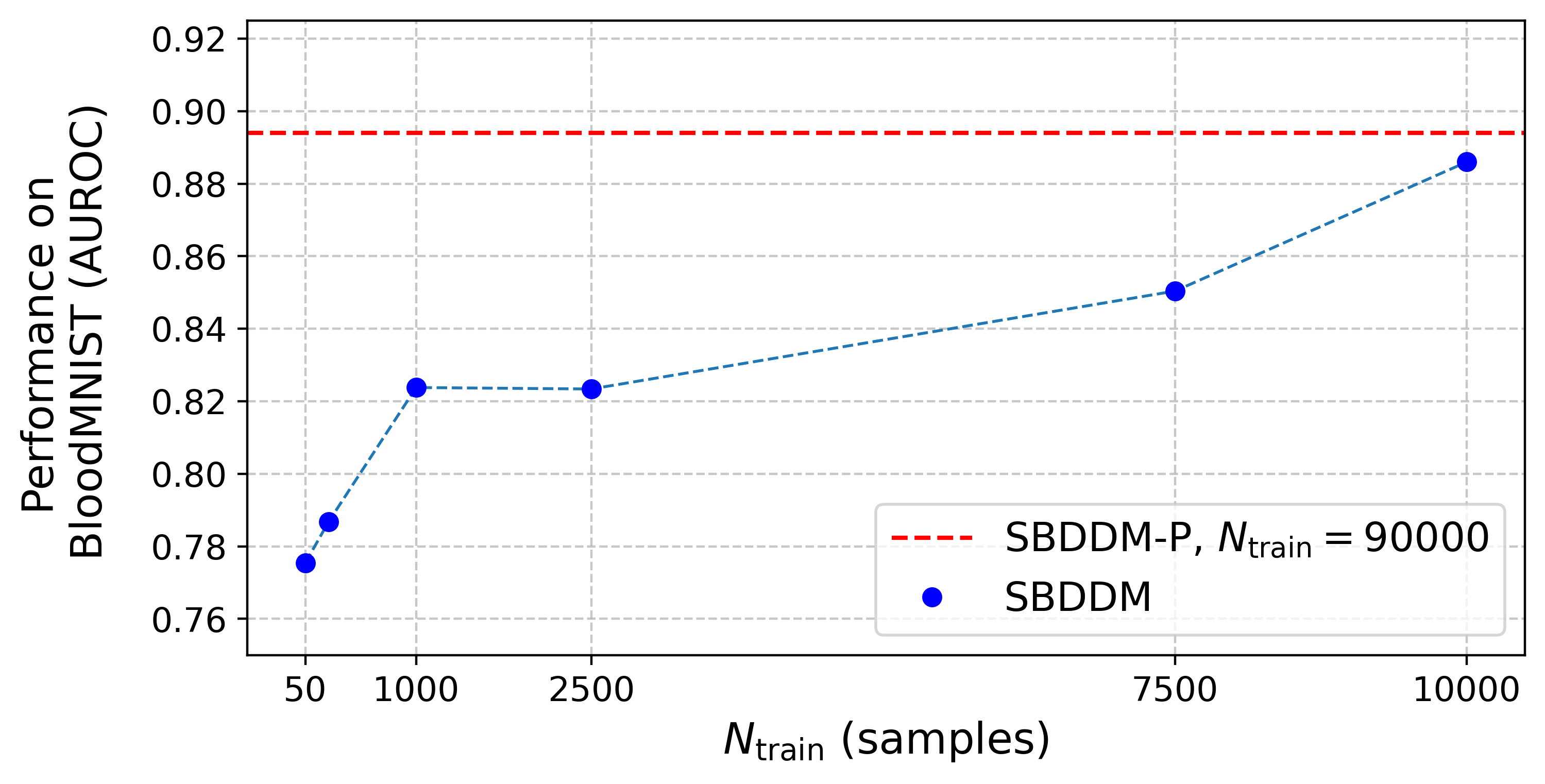}
    \caption{When is retraining worth it? Performance on the BloodMNIST benchmark using SBDDMs trained on progressively larger subsets of 'healthy' images, compared against a SBDDM-P pre-trained on 90k PathMNIST samples.}
    \label{fig:ablation2}
\end{figure}
Although score-based DDMs trained on one dataset can generalize to others, we examine under what conditions retraining on the target inlier dataset becomes worthwhile. Specifically, we empirically show how much data is needed for a SBDMM to match the performance of a SBDDM-P model. \Cref{fig:ablation2} presents performance on the BloodMNIST Near-OOD benchmark using SBDDMs trained on increasing subsets of 'healthy' BloodMNIST samples, compared to a SBDDM-P pre-trained on PathMNIST (approximately 90k samples). Results show that the SBDDM-P consistently outperforms SBDMMs trained on fewer than 10k samples. Only when the SBDDM model is trained on $\sim$10k samples does its performance begin to approach that of SBDDM-P. These findings highlight the proposed method's practicality under data-scarce conditions.

%% file: chapters/5-Conclusion.tex
\section{Conclusion}

OOD detection is critical for medical imaging systems that must handle rare or unexpected anomalies without explicit supervision. We presented an efficient diffusion model for OOD detection in medical imaging, leveraging forward diffusion trajectories from score-based DDMs. Unlike prior DDMs, our method needs no class conditioning or costly reconstruction, using only five noising steps. Exploiting the Stein score captures both semantic and modality shifts between inlier and anomalous samples. Our experiments show that a single SBDDM pre-trained on a large homogeneous medical dataset generalizes well across modalities and achieves strong performance across all benchmarks, requiring no fine-tuning beyond fitting a lightweight KDE.
 In contrast, SBDDM models trained on large heterogeneous or natural image datasets yield lower performance, highlighting the importance of domain-consistent training for robust score estimation. These findings highlight the potential of SBDDMs as practical, generalizable OOD detectors for clinical applications. Future work may explore extending this approach to volumetric imaging, semantic segmentation, and integrating uncertainty quantification to further enhance decision-making in real-world clinical settings. Additionally, while our method focuses on image-level OOD detection, it does not provide spatial localization of anomalies---a key advantage of DDM-based approaches. Incorporating localization into this framework is an important step toward interpretability in clinical workflows.